\documentclass{article}

\usepackage[nonatbib, final]{ap_2020}

\usepackage[utf8]{inputenc} % allow utf-8 input
\usepackage[T1]{fontenc}    % use 8-bit T1 fonts
\usepackage{hyperref}       % hyperlinks
\usepackage{url}            % simple URL typesetting
\usepackage{booktabs}       % professional-quality tables
\usepackage{amsfonts}       % blackboard math symbols
\usepackage{nicefrac}       % compact symbols for 1/2, etc.
\usepackage{microtype}      % microtypography
\usepackage{import}
\usepackage{dirtytalk}
\usepackage[colorinlistoftodos]{todonotes}
\usepackage{caption}
\usepackage{subcaption}
\usepackage[super]{nth}

\title{Gunrock 2.0: A User Adaptive Social Conversational System }

\author{
Kai-Hui Liang\thanks{kaliang@ucdavis.edu} , Austin Chau, Yu Li, Xueyuan Lu, Dian Yu\thanks{dianyu@ucdavis.edu}, \\ \textbf{Mingyang Zhou, Ishan Jain, Sam Davidson\textsuperscript{$\diamond$}, Josh Arnold, Minh Nguyen, Zhou Yu\thanks{joyu@ucdavis.edu}} \\
 Department of Computer Science\\
 \textsuperscript{$\diamond$}Department of Linguistics\\
 University of California, Davis\\
 Davis, CA 95616
}

\begin{document}

\maketitle

\begin{abstract}
Gunrock 2.0 is built on top of Gunrock with an emphasis on user adaptation. Gunrock 2.0 combines various neural natural language understanding modules, including named entity detection, linking, and dialog act prediction, to improve user understanding. Its dialog management is a hierarchical model that handles various topics, such as movies, music, and sports. The system-level dialog manager can handle question detection, acknowledgment, error handling, and additional functions, making downstream modules much easier to design and implement. The dialog manager also adapts its topic selection to accommodate different users' profile information, such as inferred gender and personality. The generation model is a mix of templates and neural generation models. Gunrock 2.0 is able to achieve an average rating of 3.73 at its latest build from May \nth{29} to June \nth{4}.
\end{abstract}

\section{Introduction} \label{sec:intro}
%====TEMP======
% OUTLINE:
% 1. Intro about general systems  
% 2. Alexa intro
% 3. Intro about gunrock
% 4.
% ================

Protecting user privacy is an essential task which many companies are taking seriously, this means many large, clean, real conversation data, such as customer service dialogs, are not publically available.
While this protection should not be compromised, this still poses a significant challenge for the academic community in developing advanced dialog models. 
Current dialog corpora frequently used in dialog research, such as MultiWOZ\cite{budzianowski2018multiwoz} and PersuasionForGood \cite{wang2019persuasion} are all collected through role-play formats on crowdsourcing platforms. 
No matter how well the tasks are designed, these role-play datasets have limitations compared to those from real users who have real intent to interact with the dialog systems. 
Amazon has the foresight to help academia and industry solve this problem by launching the Amazon Alexa Prize for building engaging open-domain social conversational systems. 
The Amazon Alexa Prize provides a platform that attracts a large number of volunteer users with real intent to interact with the participating social conversational systems. 
% We obtained on average more than 400 conversations per day over the course of the 30 day evaluation period. In total, we collected more than 3,000,000 conversation turns throughout the entire development period, with collection ending on April 30th. 
Anyone in the United States who has an Alexa powered device, such as Amazon Echo, can interact with our system. This platform provides us an extensive and detailed dataset for developing a large-scale conversational system and conducting dialog system research.

%As humans are accustomed to the communication patterns of one another, most users would likely transfer their human-human communicative behavioral patterns and expectations to interactions with a system. For example, while users quickly learned that Microsoft Cortana (a personal assistant) could not handle social content, 30\% of the total user utterances addressing it consisted of social content \cite{jiang2015automatic}. Therefore, one possible way to improve conversational system performance is to imitate human communicative behaviors. We propose an open-domain social bot, Gunrock, to imitate natural human-human conversations with the ability to cover a wide variety of social topics that can converse in depth on specific and popular subjects.  %The ultimate goal of Gunrock is to engage users in open-domain social conversations for longer than twenty minutes. The system is also assessed by individual users feedback on how likely they would like to talk to the system again on a 1-5 Liker scale. In order to engage users, the system not only has to be coherent but also interesting to talk to.  We proposed a hybrid system design to combine both rule-based and statistical machine learning models to increase the breadth and depth of the conversations. 
We built Gunrock 2.0 based on our 2018 Alexa Prize-winning bot, Gunrock. Gunrock 2.0 extends the design idea of user adaptation.  We made a number of contributions in open domain spoken language understanding, dialog management, and language generation.  

We improved on several aspects of the previous Gunrock system: 
1) Better understanding modules, including more fine-grained intent detection, and better-named entity detection and linking models; 
2) A generation based acknowledgment model; 
3) A user-adaptive dialog manager that controls topic selection; 
4) A generation based follow-up request and response.

% Zhou
\section{Related Work} \label{related_works}
Open-domain chatbots, such as the classic example Alice \cite{alice}, aim to pass the Turing Test. In comparison, social chatbots require more in-depth communication skills with emotional support\cite{xiaoice}. Gunrock 2.0 extends Gunrock and utilizes state-of-the-art practices in both domains and emphasizes dynamic user adaptation conversations.

Many neural models \cite{neuralconversationmodel} and reinforcement learning models \cite{rfconversationalmodel} have been proposed for end-to-end dialog generation and understanding. With large datasets available, including Cornell Movie Dialogs \cite{cornell_movie_data} and Reddit~\footnote{\url{https://www.reddit.com/}}, these models improve dialog performance using an end-to-end approach. Pre-training based dialog models such as ARDM \cite{ardm} or DialogGPT \cite{dialogpt} have been very popular as well. However, these methods still suffer from incoherent and generic responses \cite{incoherent}.  

In order to solve those problems, some research has combined rule-based and end-to-end approaches \cite{alquist}. Other relevant work leverages individual mini-skills and knowledge graphs \cite{soundingboard}. This combination of approaches enhances user experience and prolongs conversations. However, they are not flexible in adapting to new domains and cannot handle robustly opinion related requests. For example, in the 2017 Alexa Prize~\footnote{\url{https://developer.amazon.com/alexaprize/2017-alexa-prize}}, Sounding Board \cite{soundingboard} reported the highest weekly average feedback rating (one to five stars, indicating how likely the user is willing to talk to the system again) with a $3.37$ across all conversations. For comparison, in 2018, our chatbot Gunrock reached 3.62 across all conversations. 

Gunrock 2.0 takes full advantage of better understanding models, including better entity detection and linking, detailed intent detection, better acknowledgment, better user adaptation in dialog management and a generation model to handle out of domain follow-up questions. These novel concepts contributed to our rating of $3.73$  from May \nth{29} to June \nth{4} using our latest build.

% During the last week of the semifinals (August 8-15, 2018), we reached the weekly average rating of $3.56$. % Zhou
\section{Architecture} \label{sec:architecture}

We leveraged the Amazon Conversational Bot Toolkit (Cobot) \cite{ch2018advancing} to build the system architecture. The toolkit provides a zero-effort scaling framework, allowing developers to focus on building a user-friendly bot. The event-driven system is implemented on top of the Alexa Skill Kit (ASK) with AWS Lambda function\footnote{\url{http://aws.amazon.com/lambda}} as an endpoint. Each conversation turn is sent as an event request to the system. The Cobot framework uses a state manager interface that stores both user attributes and dialog state data to DynamoDB\footnote{\url{https://aws.amazon.com/dynamodb/}}. We also utilized Redis\footnote{\url{https://redis.io}} and DynamoDB to build the internal system's knowledge base.

We will cover each system component in this section.

\subsection{System Overview} \label{ssec:sys}
\begin{figure}[ht]
\includegraphics[width=1.0\linewidth]{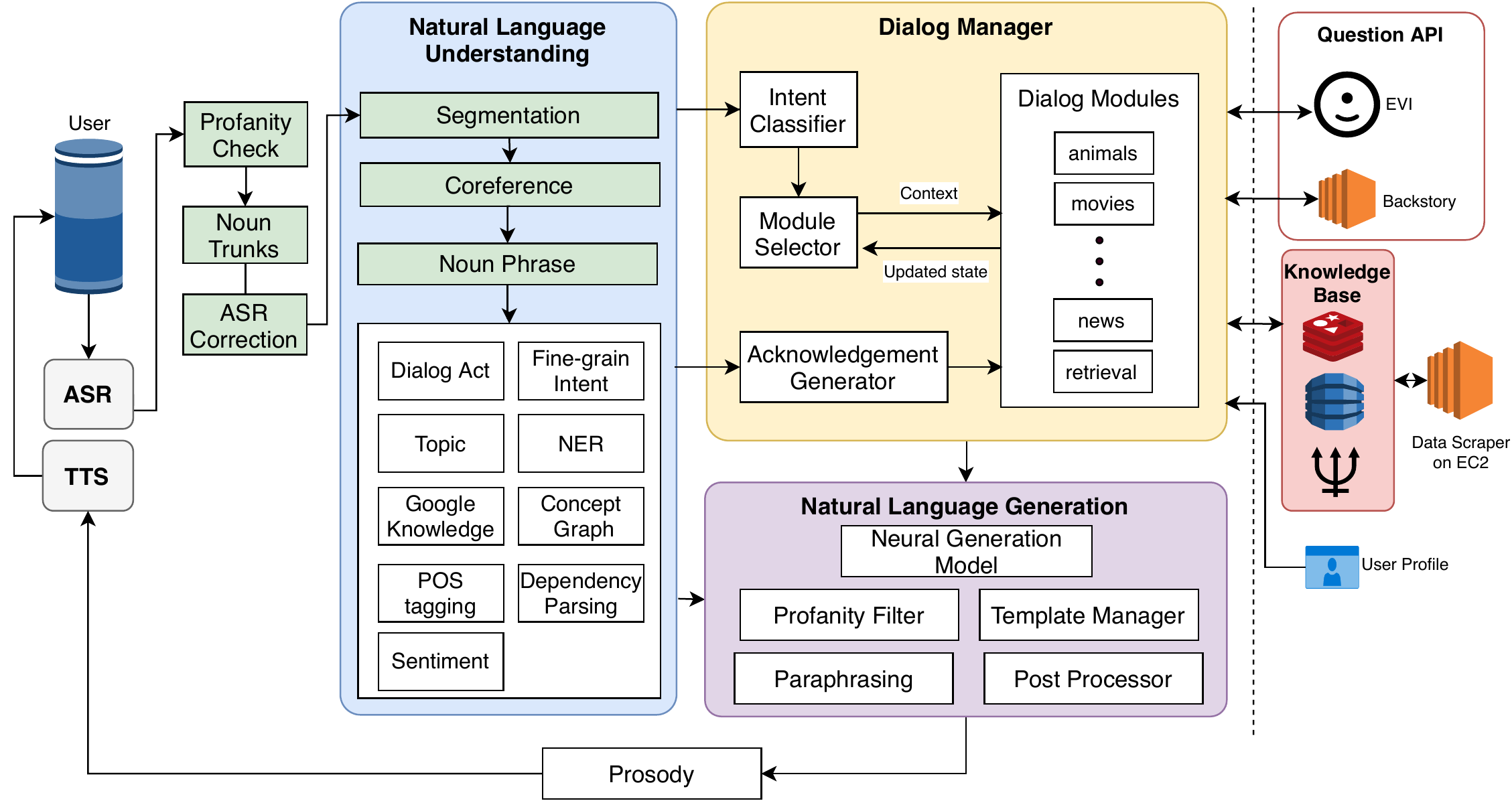}
\caption{Gunrock System Architecture}
\label{fig:architecture}
\end{figure}
Figure \ref{fig:architecture} depicts the social bot dialog system framework. Alexa Skill Kit provides the text of user utterances with timestep and confidence information through an automatic speech recognition (ASR) model. Our system takes the text input and generates text output in the Speech Synthesis Markup Language (SSML) format. Then ASK's built-in Text-To-Speech (TTS) service takes the text and generates speech output.

To avoid users talking about profane content, we detect profanity using Amazon's Offensive Speech Classifier toolkit \cite{ch2018detecting} and our custom profanity list with regex. We inform the user of the inappropriateness of the topic and recommend an alternate subject to continue the conversation. Additionally, if the ASR confidence score is low, we prompt the user to clarify their utterance. In both cases, we bypass the downstream NLU pipeline to reduce latency.

After ASR, the user input is processed by multiple NLU components. Figure \ref{fig:architecture} shows that there are $12$ components in NLU. We use Cobot's built-in natural language processing pipeline, which supports multi-threaded task execution. There are four main steps in the NLU pipeline. First, the input utterance is segmented into multiple sentences. Then we solve coreference and detect noun phrases. Finally, user utterances are analyzed by several NLP components, as shown in Figure \ref{fig:architecture}, in parallel. We will discuss the NLU components in detail in Section \ref{ssec:nlu}. 

In the Dialog Manager, the Intent Classifier combines all NLU features and further classifies them into multiple intents. Module Selector then directs different user intents to corresponding dialog modules as described in \ref{sssec:sysintent}. The dialog modules cover several specific topics, such as movies, sports, and animals. Each topic dialog module has its own dialog flow, and it adapts the flow to users' interests. We will explain this in more detail in Section \ref{sec:dialog_flow}. The dialog manager also contains an acknowledgment generator, which acknowledges the user's input depending on the context and user intents. We would describe it in Section \ref{sec:acknowledgement}.

We store our knowledge database offline on Redis and Dynamodb. We use Amazon EC2 to regularly scrape data from different sources to keep up with new content such as news and movie trivia.

% NLG  
Our NLG consists of manually curated templates, content retrieved from the internet, and responses generated by a neural generation model. For templates, we use a template manager to select templates based on the key assigned by topic modules. To ensure the appropriateness of content retrieved from the internet, we include a profanity checker to review the response's content. A post-processor is also included in NLG to modify the original topic dialog module response. Finally, we use Amazon SSML to format the prosody of the response. We will cover more details in \ref{ssec:nlg}.

% In the following subsections, we describe the major system divisions in detail. 

\subsection{Automatic Speech Recognition} \label{ssec:asr}

% Before user utterances pass through the NLU, our system prepossesses the input utterance based on the ASR overall confidence score and each word's confidence score in order to address and handle ASR errors. We define three ASR error responses based on the confidence score range:

% \begin{itemize}

%     \item \textbf{Critical Range}: If the overall confidence score and each word confidence score are below $0.1$, our system asks users to repeat or rephrase their utterance without running the downstream pipeline.

%     \item \textbf{Warning Range}: If the overall confidence score is lower than $0.4$ but is not in a Critical Range, it is allowed to pass through to ASR correction, which is discussed in Section~\ref{sssec:asr_correction}.

%     \item \textbf{Safe Range}: For other cases, we define it as a safe range and directly use the ASR result.

% \end{itemize}

Similar to Gunrock, we filter ASR outputs with a confidence threshold. The system asks the users to repeat their input instead of sending through the NLU and NLG pipeline if the ASR confidence is low.
Even though improved ASR models achieve lower word error rates, recognizing the correct words is still challenging, mainly due to new phrases and lack of context. We apply ASR correction proposed by Gunrock by comparing noun trunk phonetic information to the noun phrases in the context domain, such as movie names, using Double Metaphone \cite{double_metaphone}.
In instances where we can unambiguously capture the error phrases using regular expression, we replace the erroneous phrases with the corrected phrases before the NLU pipeline processes them. 

\subsection{Nature Language Understanding} \label{ssec:nlu}
% Cobot provides topic classification, sentiment analysis, profanity check and NER detection for NLU. We add a sentence segmentation model to separate input from users into semantic units and perform NLU on each semantic segment. After extracting noun phrases, we implemented (named) entity recognition, co-reference, ASR correction, and dialog act detection to support language understanding. We present the technical details of each model in the order of the NLU pipeline (Section \ref{ssec:sys}).
We follow the general NLU pipeline of Gunrock \cite{yu2019gunrock} to segment user input into semantic units after ASR correction and apply coreference resolution, constituency parsing, noun phrase extraction, named entity recognition, and sentiment analysis on each segment. We also add models such as sentence completion and dependency parsing for in-depth understanding and focus on improving each model with active learning strategies.

\subsubsection{Gunrock NLU} \label{sssec:np}
Our NLU pipeline is similar to Gunrock. Specifically, we borrow the coreference resolution method with error fixes, constituency parsing for noun phrase extraction, and entity linking with Google Knowledge Graph ~\footnote{\url{https://developers.google.com/knowledge-graph/}} and Microsoft Concept Graph ~\footnote{\url{https://concept.research.microsoft.com/Home/Introduction}}. We improve other models detailed below.

% \import{nlu/}{asr_correction.tex} 
\subsubsection{Sentence Segmentation} \label{sssec:np}

Gunrock converts the sentence segmentation into a machine translation problem by training a sequence to sequence (s2s) \cite{s2s}  bidirectional LSTM model \cite{lstm} using the Cornell Movie Quotes corpus \cite{cornell_movie_data}. The source is the corpus after removing all the punctuation, and the target is to recover the sentence boundaries. With masking NER as pre-processing, maintaining original tokens as post-processing, and setting a timestop threshold, the model achieves $84.43\%$ F1 score but performs worse on longer utterances. To improve the model performance, we experimented with different models, including transformer \cite{transformer} as an s2s model and a sequence tagging model to predict sentence boundaries. With more annotated in-domain data and longer context, we achieve an F1 score of $91.9\%$ on an annotated development set.
% \import{nlu/}{coreference.tex}
\subsubsection{Dialog Act Prediction} \label{sssec:dialog_act}

Gunrock uses MIDAS \cite{yu2019midas} dialog act scheme for multi-label dialog act prediction in a hierarchical structure. For more robust dialog act prediction incorporating context, we use Hybrid-EL-CMP \cite{zhang2019filling} by completing utterances and predicting dialog act with a combined distribution of the ellipsis and the automatically completed utterance.  In addition, we improve our dialog act model by using active learning strategies \cite{active_learning}. In specific, we extract utterances using regex to boost training examples for low-performance labels, and we also choose utterances with low prediction confidence. We retrain the model with an updated training corpus and keep improving model performances during deployment.
\subsubsection{Fine-Grained Intents} \label{sssec:fine_grain intents}
We define extra intents as fine-grained intents that are used to complement our dialog acts. These fine-grained intents provide supplemental labeling of the user intents for our downstream tasks. We use regex to detect these intents based on the conversation we collected throughout the competition. For example, if users ask ``\texttt{ask\_back}`` questions (e.g. ``what about you? ''), we would answer it according to the current topic. If users express their opinion, we use ``\texttt{opinion\_positive}'' (e.g. ``that's interesting'') and ``\texttt{opinion\_negative}'' (e.g. ``it's boring'') to understand user's opinion positivity. We also use ``\texttt{ans\_like}'' (e.g. ``I'm into movies'') and ``\texttt{ans\_dislike}'' (e.g. ``I'm not a big fan of movies'') to detect users' preferences towards a topic.
\subsubsection{Noun Phrase Extraction} \label{sssec:np}

Semantic parsing such as AMR \cite{amr} and UCCA \cite{ucca, dag_ucca} has been popular to facilitate language understanding tasks by extracting relationships among entities. However, they are expensive to annotate and hard to transfer to dialog domains for good performance. On the other hand, sequence tagging methods for NER detection have been well studied and are reliable with the benefit of large language models \cite{bert}, but do not work well for regular nouns, which are essential for topic detection. Therefore, we follow the constituency parsing method of Gunrock to get noun phrases in different levels in the constituency tree. 

However, even though such a method extracts noun phrases exhaustively and achieves high recall, it suffers from two problems: low precision; and lack of understanding of relationships. Therefore, we may not be able to detect the true topic with multiple noun phrases. 
To solve the first problem, we use a BiLSTM-CRF sequence tagging model to identify noun phrases that are important for understanding.
We obtain training data for this model by filtering out non-informative noun phrases from the constituency parser.
We referred to this model as the KeyPhrase model.
Since the labels in the training data are noisy, we used entropy regularization~\cite{reed2014training} during training so that the model could reject inconsistent labels.
For the second problem, we use the extended SCUD annotation scheme \cite{dependency} and build a dependency parsing model on annotated dialog data to extract relationships among entities. The extracted noun phrases are used as input to entity linking for topic detection.
\subsubsection{Entity Linking} \label{sssec:ner}

Recognizing and linking entities such as movies, books, and people are critical in a dialog system.
The problem of entity detection and linking is even more challenging in spoken settings than in written text because of the following problems. First, modeling context is harder because users can avoid questions and mention completely different things.
However, modeling context is important to resolve whether Harry Potter is a ``fictional character'' or a ``movie''.
Second, the input text is noisier, including grammatical errors, ASR errors, dis-fluency, and more.
The absence of reliable casing information also makes this a much harder task than written text.
Third, the diversity of entities is higher in open domain chit-chat than in written documents, which usually focus on a few narrow topics.
Fourth, users may recall incomplete or erroneous names of entities.

In addition to the spans of text with which could be entities, the detection model also predicts the categories of the spans.
The categories limit the scope of the linking targets, e.g., only movie names or only character names, making linking more precise.
We used an ensemble of models (experts) where each model is responsible for entity linking for a particular topic module.
We used the LSH-Ensemble~\cite{zhu2016lsh} model to ensure that linking time is short.
The text span is linked to an entity based on the similarity between the text and the name of the entity.

We also designed a similarity metric function to ensure that linking is more robust against errors in the text. For instance, Google Knowledge Graph returns the song ``Pink Fluffy Unicorns Dancing On Rainbows'' for the query ``unicorns'' and returns the singer ``Charley Pride'' for the query ``charlie pride'' with different spellings. The latter is a valid matching while the former is not. To consider the similarity in the semantic level, we use BERT \cite{bert} to encode both the query and the returned result and compare their similarity in the hidden states. In addition, we consider the context for the hidden representation. Specifically, we feed the system and user utterances as consecutive sentences to the BERT model and get the mean hidden representation of the entity tokens to represent the query.
Similarly, we concatenate the returned phrase (e.g. ``Red Dead Redemption 2'') with its description (e.g. ``Survival game'') and get the mean hidden state of all the tokens to represent the knowledge embedding. We use cosine similarity, F1, and F2 distances as the similarity metric.  Cosine similarity with a pre-defined threshold works the best in our empirical analysis.
\subsubsection{Sentiment} \label{sssec:sentiment}

Sentiment analysis is mainly used for three purposes: to detect user intent together with dialog act, to detect user engagement, and help user utterance acknowledgment explained in Section \ref{sec:acknowledgement}. Instead of the standard binary prediction, such as the IMDb dataset \cite{imdb} and SST-2 \cite{sst} or fine-grained five-class prediction, such as SST-5 \cite{sst}, three classes of positive, negative, and neutral are more critical in our system understanding. Therefore, we map SST-5 into three classes in the sentence level and train a BERT \cite{bert} model. The model reaches an accuracy of $84.38\%$.

\subsection{Dialog Management} \label{ssec:dmgnt}
We created a two-level hierarchical dialog manager to handle user inputs. The high-level system selects the best dialog module for each user request leveraging the output from NLU. After that, the low-level system activates the selected dialog module to generate a response.

% \subsubsection{High-Level System Dialog Management} \label{sssec:sysintent}
\subsubsection{High-Level System Dialog Management} \label{sssec:sysintent}
High-level dialog manager \textit{module selector} selects a dialog module to handle user utterance by detecting user's \textit{intent} and referring to the \textit{dialog module state} of the previously selected module.

\paragraph{Intent Classifier}
We define \textit{functional intents}, \textit{strong topic intents} and \textit{moderate topic intents} to classify user utterances based on the human-bot conversations we collected during the competition. We break the user utterances into smaller segments, and each segment may have multiple intents.

\begin{itemize}
    \item \textbf{Functional Intents}:
    \begin{itemize}
        \item \textbf{Incomplete and Hesitant Utterances}:
            Due to endpointing model accuracy constraints in the Alexa ASR system, sometimes user utterances may be incompletely captured when a user has paused. To address this issue, we ask users to repeat when their utterances are detected as incomplete (eg. ``I think it's'') or hesitant (eg. ``let me think''). 
        \item \textbf{Clarification Intent}:
            If a user asks the bot to repeat (eg. ``what did you say''), the system repeats its last response.
        \item \textbf{Device Task Request}: 
            Our system detects Alexa device system requests (eg.``volume up'') by guiding users to exit social mode to enable device functions. If the request is similar to a topic our system can handle, (eg. ``play songs by Taylor Swift''), the system will ask the user if they are interested in talking about Taylor Swift.
    \end{itemize}

    \item \textbf{Strong Topic Intents}:
        \begin{itemize}
        \item \textbf{Topic Switch Intent}: 
            If the user explicitly expresses their disinterest in the current topic (e.g. ``I don't want to talk about movies anymore'') or wants to change a topic (e.g. ``can we talk about something else''), the system proposes another topic or asks what topic user wants to talk about.
        \item \textbf{Topic Request Intent}: 
            If the user specifies a topic with a command, e.g. ``let's talk about movies'', the dialog module selector immediately selects the movie module.
        \end{itemize}

    \item \textbf{Moderate Topic Intents}:
     \begin{itemize}
        \item \textbf{Topic Preference Intent}: 
            Our system detects users' preference for a proposed topic to decide whether we should continue talking about it. For example, if system asks ``Do you like to watch movies?'', and user replies ``I'm not a big fan of movies'', the system will not continue the proposed topic even if user doesn't explicitly reject it by saying ``no''.
        \item \textbf{Topic Intent}:
            Our system has a hierarchical method to detect topic intents in user utterances. First, based on all the dialogs we collected, we created a database to map user utterance to our topics. Since this database is human-annotated, it has a lower chance of false-positive topic detection. Thus, we use it as the first level to detect topics. In case we do not detect topics in the first level, we combine Google Knowledge Graph, Microsoft Concept Graph, and Cobot's built-in joint Dialog Act/Topic Model as our second level to detect topics. 
            We tune the confidence thresholds of these three detectors based on the heuristic and dialog data we collected. One disadvantage of this second level is that it detects many false positives in the utterance as movie names, book titles, and song names. To avoid such false positives, we do not map detected movie names, book titles, and song names to topic modules.
    \end{itemize}
\end{itemize}

\paragraph{Dialog Module State}
The previously selected module would be in one of the three state: ``\texttt{CONTINUE}'', ``\texttt{UNCLEAR}'' and ``\texttt{STOP}''. If the module has not finished its conversation, it will set itself to ``\texttt{CONTINUE}'' state, and the module selector will select it again if no functional intents or strong topic intents are detected.
If a module detects that the user is talking about other topics, the topic module will attempt to propose another module that can handle the user's interests. It confirms whether the user wants to switch topics and sets itself as ``\texttt{UNCLEAR}'' state. If the user agrees, then the system switches the topic. If no clear intent or no other topics are detected, the system will select the same module again. When the conversation reaches a point where the current module has exhausted most of its predefined topics, the module will set ``\texttt{STOP}'' state to stop the current topic, and then the system proposes the next topic in the same turn.

\paragraph{Dialog Module Selector}
To select the most appropriate module, we set priorities for the detected intents. For functional intents, we select a functional module. For strong topic intents, we switch to the corresponding topic module no matter which states the previous module is in. Otherwise, we either select the previous module or propose a new topic depending on the state of the previous module.
When the previous module is in ``\texttt{UNCLEAR}'' or ``\texttt{STOP}'' state and multiple topic intents are detected, if one of them is the same as the previously selected topic, we prefer to select this topic over others. Otherwise, we would propose the topic with the highest priority based on the personal topic order described in \ref{sec:adaptation}. In the next turn, if the user accepts the topic, our selector would select the proposed module.

\paragraph{Error Handling}
To handle system or dialog module crashes, we designed an error handling method based on the dialog act and fine-grain intents to keep the user engaged. For example, if the dialog module crashes when we detect fine-grain intents as ``ans\_like'', we would say ``Awesome, I like that too. I guess we have similar tastes'' to continue the conversation. If the crash happens at the system level without any available NLU feature, the bot would say ``My bad, I lost my train of thought. Do you want to try again, or would you rather talk about something else? '' This way, users know something went wrong on our side, and it also gives users a chance to change the topic. In addition, in our user studies, we noticed that when the users give long utterances while the internet is not very stable, the Alexa Skill Kit server fails to get user input events and thus triggers a reprompt. Our bot sets reprompt template as ``Thanks for sharing, Your thoughts are really interesting, but I’m struggling to keep up. Can you explain that again more simply for me? '' to guide the user to speak a shorter utterance to avoid the same issue happening again.

% \subsubsection{Low Level Dialog Management} \label{ssec:modules}
\subsubsection{Low Level Dialog Management} \label{ssec:modules}
The Low-level dialog management consists of several dialog modules. There are two types of dialog modules. \textit{Functional dialog module} handles functional intents such as device task requests. Each \textit{Topic dialog module} handles one major topic, such as movies, music, and sports. Each topic dialog module has its own dialog flow design and uses \textit{Finite State Machine} to manage dialog flow. 

\paragraph{Topic Dialog Modules}
 We keep the most popular topic modules from Gunrock, including movies, books, music, games, animals, sports, food, travel, news, technology and science, and retrieval. We improved their dialog design and extend the content as described in Section \ref{sec:dialog_flow}. We also add new modules users are interested in, for example:
 \begin{itemize}
    \item \textbf{Daily life and Outdoor Activities}: These two modules handle high-frequency topics users mention when answering open questions like ``What do you like to do in your free time? ''. Daily life covers topics in users' everyday lives, including friendship, family, relationships, playing with mobile phones, and chores. Outdoor activities covers hobbies such as biking, hiking, and kayaking.  We start with a positive acknowledgment to increase the engagement (e.g., ``It's really nice that you spend time playing with your kids. You must be an awesome parent!'').  At the end of the topic discussion, we propose a relevant topic module to make the transition more natural and coherent. (e.g., ``do you have a family pet? '').
    
    \item \textbf{Fashion}: The Fashion module discusses niche topics, including makeup, shopping for clothes, and fragrances. User adaptation to gender and agreeableness is implemented to propose more engaging sub-topics and respond to the user's interest level. The Fashion module was developed primarily for feminine audiences, while still containing neutral topics to adapt to a broader audience. In addition, a recommendation feature was developed that elicits user preferences to match them with an appropriate fragrance based upon their likes/dislikes. More detail is described in Section \ref{ssec:adaptation_in_topic}.
    
    % \item \textbf{Fashion}: The fashion module covers topics like fashion brands, makeup, clothes, and fragrances. It adapt subtopic based on user's user profile and agreeableness as described in \ref{ssec:adaptation_in_topic}.
    
    \item \textbf{Comfort}: The comfort module is selected when users reveal they have had a terrible day or are having a hard time. We design special acknowledgment and validation for various conditions such as loneliness, worries, and sickness. After listening to the user's feelings and experience for several turns, it then proposes funny content like telling jokes to cheer the user up.
\end{itemize}

\paragraph{Finite State Machine (FSM)}
Our topic modules use a finite state machine (FSM) to represent each dialog state and handle user responses. We choose to use a finite state machine because it allows greater control of the dialog flow so that the focus of engineering and experimentation can be on the dialog design and upstream NLU pipelines.

We create a custom FSM manager that provides the machine for state transition and a lightweight class structure for designing the modules' dialog flow. 
The FSM manager is state-centric, meaning transitions are defined within states. This design is chosen since traditional top-down approaches in designing FSMs are inflexible and hard to modify for our usage. Dialogs are dynamic and varied, and we emphasize acknowledging users' responses, which would be hard to encode using traditional methods such as transition tables.

Each state is identified by a unique string, which is used to define each dialog transition. A state also encapsulates all or part of the logic that should be executed in each dialog turn. Thus, each state can be loosely defined to represent a complex dialog turn, or part of the utterance depending on the clarity and reusability for each dialog flow. It can make API requests, generate template responses, or decide the next state transition based on the context. 
Each state can signal the FSM manager to perform a transition by returning the target state's identifier, and whether such a transition should happen in the current turn or the next. If the transition happens within the current turn, the system can chain states to create more complex responses. These transitions are defined inside the state because other logic, such as template generation and API queries, are defined within the state. Grouping them makes each state the only source of what should happen during a dialog turn. 
We find this makes designing our dialog flow more efficient and allows the reuse of parts of an utterance for similar responses. We also find defining the next state transition alongside the logic is more scalable than separating them since we only need to refer to each state to inspect the system's decision at each turn.

For each state, a tracker object is used to wrap around stateful storage and upstream NLU components, which can be passed between states and assign variables within the dialog turn or throughout the conversation. This tracker object allows us to encapsulate the upstream NLU pipelines and contextual information for user adaptation. 
Since the dialog flow is highly dependent on the upstream NLU results, this layer provides flexibility in our implementation to change and scale either efficiently.

\textbf{Question Handling}
All the topic dialog modules use a common question handler to handle general questions. The question handler uses our custom back-story database to answer questions associated with our bot's persona, such as \textit{What is your favorite color?}, and uses Amazon's EVI service\footnote{\url{https://www.evi.com/}} to handle factual questions. If there are no answers from both of them, we use our acknowledgment generator to generate a specific rephrased answer to express that the bot does not have an answer as described in \ref{sec:acknowledgement}. If the acknowledgment generator fails to generate a response, we then generate a different response based on the dialog act. For example, if the dialog act is ``\texttt{OPEN QUESTION OPINION}'', we then reply with ``Good question, I haven't thought about that before. '' to maintain an engaging conversation even we do not have an answer.

% Maybe talk about this in the next version?
% \textbf{Multiple Segments Handling}:
% (we check all segments instead of just check one segment to avoid missing information. Describe how we handle different segments combinations)

\subsection{Natural Language Generation (NLG)} \label{ssec:nlg}

Our NLG module is a mixture of template-based and neural generation methods. It selects a manually designed template and fills out specific slots with information retrieved from the knowledge base by the dialog manager. It uses neural generation for unfamiliar topics and follow-up requests. 
% The template manager module avoids response repetition and generates utterances with a variety of surface forms.
We also use Amazon Speech Synthesis Markup Language (SSML) to provide our generated responses with prosodic variations.

%which provides APIs for the rest of the system to generate responses.

\subsubsection{Template Manager} \label{sssec:tm}
The template manager centralizes all response templates for our system. Its design is similar to Gunrock. By consolidating all templates, we can avoid having duplicate responses. Each response is identified by a unique key, under which are templates of varying surface forms to make our content more natural and human-like. We use a shuffle bag method to ensure no templates are repeated after another.

Each template may also contain slots for dynamic substitution. This setup allows the insertion of retrieved data, such as weather or metadata of a movie. It also allows the insertion of captured NER from the user's response when generating acknowledgment.

We have expanded the capability of the template manager by allowing key-values to be encoded alongside each template. This allows for more information to be retrieved per template usage. For instance, a musician specific template can be stored with key-value pairs describing the musician, its identifier in our knowledge base, and a name that is formatted for the speech synthesis. This grouping allows for better efficiency when using the template response.

\subsubsection{Neural Generation Model} \label{sssec: generation}
We believe that a dialog system, unlike humans, should be able to handle any topic when it has access to a large database or internet instead of responding generic utterances or changing to familiar topics. Therefore, we designed a retrieval module to retrieve relevant information from online sources similar to Gunrock \cite{gunrock, yu2019gunrock}. However, the retrieval method is not robust with handling follow-up questions and comments. Therefore, we need a neural generation model to have a conversation about the retrieved information.

Text generation models are known for problems, including generating repetitive and non-specific sentences \cite{non-specific}. Recent large pre-trained language models such as GPT-2 \cite{gpt2} and models fine-tuned for dialog domain such as DialoGPT \cite{dialogpt} and Meena \cite{meena} achieve good performances on open-domain dialog. However, they still face the challenges of storing all knowledge into the model weights, thus cannot generate reasonable responses for unfamiliar topics, and the generated text highly depends on the training data. In addition, the generation process is not controllable. Another line of research is to collect dialog dataset based on pre-defined knowledge such as Persona-Chat \cite{persona}, Wizard-of-Wikipedia (WoW) \cite{wizard_of_wiki}, and Topical-Chat \cite{topical_chat}. Models trained on these datasets are potentially more controllable by conditioning on specific attributes, but due to the nature of data collection, the collected utterances strictly follow the given knowledge and become less natural in conversation. More importantly, they rely heavily on the quality and relevance of the given attributes. To utilize the best of both worlds, we propose to inject knowledge into the pre-trained large language model. Specifically, we use DialoGPT-medium as our base model for latency concerns. We finetune the base model on the WoW dataset with KL-divergence. Similar to our retrieval method, we use Reddit as our database and use the retrieved knowledge as the condition if there is relevant information. The model learns to copy the knowledge and paraphrase the information. For instance, given the knowledge, ``Covid-19 has infected more than 1.85 million people and killed at least 114,000 worldwide, according to Johns Hopkins University." and the context ``... USR: I'm pretty good. Just a little bit worried about the coronavirus recently" in part of the conversation, the model generates "Covid-19 is a virus that has infected more than 1.85 million people, it scares me a lot" which rephrases the information correctly.

Even though DialoGPT achieves good performance on open-domain dialog, the responses are not conversational and may generate ungrammatical sentences due to its training data. Therefore, we also deploy the Blender \cite{blender} encoder-decoder model with 2.7B parameters trained with collected human conversational data \cite{persona, ed, wizard_of_wiki} for more engaging response generation. Despite that it may still hallucinate knowledge \cite{blender}, we mainly use this generation model for two purposes, question handling (such as responding to the user utterance ``what about you what do you think about sukiyaki") and unfamiliar topic handling including follow up comments (such as responding ``I love candy too! I love all kinds of candy, especially chocolate" to the user utterance ``I like talking about candy I like candy").
In specific, because of the large number of layers in the model and the auto-regressive nature of generation, we use a beam size of one and a minimum beam length of five and ten, respectively, for question handling and unfamiliar topic handling to reduce the latency to a reasonable scope (less than one second end to end). In addition, since the training data is human-human conversations, we apply a rule-based filter to clean utterances that demonstrate a human personality such as ``I like to eat at home with my wife", which clearly contradicts the personality of the Alexa social bot.
Moreover, if the generated response is truncated due to latency issues or has an undesirable dialog act, we use the original question handler and retrieval responses instead.

% However, if the retrieved information is similar in the next turn or less relevant to the context, the model may repeat the same knowledge or generate incoherent responses. In addition, even with KL-divergence, the response is less conversational. Therefore, we propose another method to do weighted-decoding so that we can shift the generation distribution depending on the knowledge while maintaining the original language model distribution. We expand the prediction probability of the next token in the language model using Bayes Rule and train the model to learn a likelihood of the knowledge conditioned on the generated token candidates while freezing the prior distribution. We then sample tokens from the new distribution to generate a response. This model is more robust towards the quality of the retrieved information. Even with less power in copying exact tokens, it can maintain the semantic meanings of the knowledge.

\subsubsection{Prosody Synthesis} \label{sssec:prosody}
% SSML
Based on the analysis of adjusting speech synthesis to make our response more human-like \cite{yu2019gunrock}, we utilize Amazon SSML\footnote{\url{https://developer.amazon.com/en-US/docs/alexa/custom-skills/speech-synthesis-markup-language-ssml-reference.html}} to format our responses before TTS. Specifically, we follow Gunrock to add fillers, insert pauses, and change speech speed based on the context of the utterance.

 % overview of the architecture, Kaihui, Dian, Minh
% \import{./}{entitydetectionlinking.tex}
\section{Dialog Flow} \label{sec:dialog_flow}

\subsection{Improved Dialog Flow} \label{improved_dialog_flow}
In order to create an engaging conversational experience for the user, we have improved our dialog flow design both on the system level and topic module level.

\subsubsection{System Level Dialog Flow} \label{improved_dialog_flow}

\paragraph{Topic Proposal and Transition}
When the current topic conversation comes to an end, the next topic can be proposed either by the current dialog module or the system. To make a seamless transition between modules, we set ``propose\_topic'' (corresponding to dialog modules) and ``propose\_keywords'' (keywords users mention) as global attributes to enable module-system and module-module communication. 

If the current dialog module detects keywords that might fit another topic module better (e.g. when user mentions "corona virus''), it sets state to ``\texttt{UNCLEAR}'',  ``propose\_topic'' to ``\texttt{NEWS}'',``propose\_keywords'' to ``corona virus'', and confirms with the user if (s)he wants to talk more about corona virus. If the user agrees, the News module then seamlessly follows up on the specified keywords at the next turn. 

The current module can also propose the closest topic after it finishes its sub-topic dialog flow. For example, after discussing users' opinions about the football team ``Seattle Seahawks'',  Sport module can send the topic ``Seattle Seahawks'' to News to continue on the topic.
% set state to ``\texttt{STOP}'', ``\code{propose\_topic}'' to ``\texttt{NEWS}'', and ``\code{propose\_keywords}'' to ``Seattle Seahawks'' and News module can follow up by providing news about the Seattle Seahawks. 

If current dialog module set state to ``\texttt{STOP}'' without setting ``propose\_topic'', the system then proposes the next topic as described in Section \ref{sec:adaptation}.

\subsubsection{Topic Module Dialog Flow} \label{improved_dialog_flow}
\paragraph{Movie}
From our user studies, we realize most users enjoy the movie trivia we provide, but some do not like how we repeatedly ask them to think of a movie to talk about. To address this, we propose a movie if the user cannot give an input or whenever we finish talking about a movie. The proposed movie is either a popular movie or a movie similar to the previously discussed one if they have common IMDB keywords (e.g., ``space'', ``family'', etc.) By interleaving between asking and proposing movies, we reduce user's cognitive load in the conversation and provide useful movie recommendations that users might be interested to know.

\paragraph{Music}
Music module in last year's Gunrock was mainly focused on pop music and singers. From the open question that we ask in system-level (\ref{ssec:open_question}), we find that many users enjoy playing instruments, especially piano and guitar. We design flows discussing users' instrument learning experiences and recommends popular instrument artists to the users.

\paragraph{Games} We have included a subtopic for discussing certain popular games in depth. For instance, if the user wants to talk about a particular game with a subtopic, we can discuss these games in-depth with carefully designed templates and questions. For example, if the user wants to talk about Animal Crossing, we can discuss specific information, such as costs of house upgrades, that would interest the user. We also adjust the module to recommend selected games at the beginning or when the user has no preference in any of the subtopics. This allows us better control of the dialog flow while providing in-depth content that crafted templates can offer. Finally, if the user mentions a game that we do not have information or templates for, we ask the user to provide a free-form description of the game. We then use different acknowledgments to show the system understands the user's response. This allows us to handle unknown entities while giving users a sense of control of the dialog.

\paragraph{Food} Built on top of food modules from Gunrock, we redesign the dialog flow to further adapt to user's interest in different subtopics in food. Our bot can talk about cooking, cuisines, and specific dishes or food that users like with carefully designed topic questions for each subtopic. When users do not provide specific interest to any subtopic, we introduce subtopics in the order of general interest to the audience in the food domain. For example, we initialize the conversation with questions like ``what's your favorite dish?'' that are more relevant to the majority of users and then gradually propose topics like ``growing a garden'' and ``healthy eating'' that are likely of interest to a smaller group of people.

\paragraph{News}
We revamped a large part of the news module this year to include more news content, and improved dialog flows to handle this content. We summarize the news articles and present them in multiple turns to better detect user intent and interest. We also include separate dialog flows for trending topics like coronavirus before presenting news about it to help make the flow more natural, engaging, and fluid. Also, we include more dialog flows containing debatable topics, which leads to a more engaging dialog.  
%Apart from that, we leverage Named Entity Recognition, Google knowledge graph to identify entities in user’s utterance to find relevant and most recent news about those entities. 
We experiment with a rule-based question generation model to ask the user if they want to continue with the news by generating questions for the next sentence in the news article. This ensures that the user is more informed about what will come next in the article before deciding if they wish to continue.

% \paragraph{Jokes}: We found users enjoy interesting and funny content, so we extend more jokes and lengthen the conversation.

\subsection{Extended content}
Providing interesting content is always a priority for us to improve user experiences with our bot. This year, we continue to spend a great effort to collect high-quality web information to discuss with our users in different sub-modules. 
\paragraph{People Also Ask}
People Also Ask \footnote{\url{https://www.internetmarketingninjas.com/blog/search-engine-optimization/googles-people-also-ask-related-questions/}} is a function from Google which provides a series of questions related to the search term that users have queried. Along with the list of these questions are the summarized answers that Google has extracted from different web pages. This function allows us to provide some insights to the topic word that people are interested in. We scrape the People Also Ask questions along with their answers for several popular topic words in the Food module as the first step to evaluate this data resource. If a user indicates interest in these topic words when talking to our food module, for example, ``I like eating steaks,'' our bot will use the content from People Also Ask to provide interesting facts about the topic word. Instead of directly presenting a fact about the topic word, we create a more interactive way by asking if the user is interested to know the answer for a certain question from People Also Ask, such as ``what is the best steak restaurant in the united states? '' and ``Is steak good for health? ''. If users are interested, we present the answer associated with the question to the user. We use this function as a point of reference when designing the dialog flow for our sub-modules.
%We are still evaluating people's degree of engagement with the content we provide from People Also Ask. 
%In the future we plan to construct a data source from People Also Ask for a massive amount of topic words across different modules such that all the modules can provide high quality facts to users. 

\paragraph{I Side With}
I Side With \footnote{\url{https://www.isidewith.com/polls}} is a website that allows people to vote on various debatable topics and share their opinions about them. We use these topics, content, and polls to create engaging dialogue flows in the News module sharing opinions from both sides of the debate while acknowledging and respecting user's personal opinions. We collect top pro and con opinions for the topic to present to the user.
\section{Adaptation} \label{sec:adaptation}
\subsection{User Profile} \label{ssec:user_profile}

Our system keeps track of the user's profile and stores static data like user name, predicted gender, and dynamic data, including preferred topics and conversation style. The user name is stored when we greet the user by asking their name. Given the name, we predict user's gender based on a popular name database from U.S. Social Security Administration (SSA)\footnote{\url{https://www.ssa.gov/oact/babynames}}, which provides baby names from 1980 to 2018 with gender information.  The preferred topic is collected when users reveal their interests when answering the open questions described in \ref{ssec:open_question}. To learn users' conversation style, we dynamically update and store the user's dominant turn ratio described below.

\subsection{Open Questions} \label{ssec:open_question}
An important goal of our system is to increase user engagement by finding the topics that users are interested in. Previous systems start a conversation by proposing one or several topics directly. It is rigid, unnatural and may not adapt well to different users' interests. In our system, we propose natural open questions when we start a new conversation to attract the user's attention as well as to know more about the user's interests. Based on the user's response, we then select the closest topic if possible.

\textbf{Open Question Categories}: 
We design three types of open questions: ``\texttt{ASK\_HOBBY}'', ``\texttt{PAST\_EVENT}'' and ``\texttt{FUTURE\_EVENT}''. ``\texttt{ASK\_HOBBY}'' is a batch of open questions that ask users' hobbies (e.g., ``What do you like to do for fun?''). ``\texttt{PAST\_EVENT}'' is a batch of open questions that ask users' past experiences (e.g., ``What was the last thing that made you smile?''). ``\texttt{FUTURE\_EVENT}'' is a batch of open questions that ask users' future plans (e.g., ``What are your plans for the weekend?''). We leverage these natural open questions to start the conversation to get users' preferred topics.

\textbf{Open Question Handling}:
When the user answers open questions, we use the module selecting strategy in Section \ref{sssec:sysintent} to select a topic module to handle it. If the system does not have an appropriate topic module, we will acknowledge the user's answer, as explained in Section \ref{sec:acknowledgement}, then naturally propose a topic for the user. On the other hand, if the user mentions multiple topics that our system can talk about, the system will save these topics as \texttt{preferred\_topics} in the user profile. When the system is about to propose the next topic, these topics are given higher priority since the user already revealed interests in them.

\textbf{Open Question vs. Topic Proposal}:
Since sometimes users do not mention all their interests in the first open question, our system will ask different types of open question after the user finishes talking about a topic. This way, we get more chances to know about the user's interest and select the corresponding topic accordingly. If the system already collected more than one \texttt{preferred\_topics}, it will not ask the next open question until all the preferred topics are discussed.

\textbf{Submissive Users vs. Dominant Users}:
We observed different types of user conversation styles in our system. Submissive users tend to follow the dialog flow initiated by the system, whereas dominant users like to direct the conversation and are more willing to express their opinions. Based on this observation, we have different strategies to fit their conversation style. Specifically, we define dominant turn ratio, calculating how many turns user utterance contains  ``\texttt{QUESTION}'', ``\texttt{COMMAND}'', ``\texttt{OPINION}''and  ``\texttt{STATEMENT}'' dialog acts over all turns. If the ratio is higher than a threshold, the user is classified as a dominant user. We ask them open questions during topic transition since they are more willing to express their thoughts. On the other hand, we directly propose topics for submissive users as the open question may give them a high cognitive load and decrease their engagement. The ratio is updated every turn, and thus, it can adapt to the user's current behavior. 

\subsection{Next Module Selection} \label{ssec:module_selection}
When the system finishes a topic and transits to the next module, our module selector takes three topic lists into consideration:
\begin{itemize}
    \item \textbf{\texttt{preferred\_topics}}: A list of all the topics that the user mentions when answering the system's open questions. These topics are likely to attract users' attention.
    \item \textbf{\texttt{personal\_topics}}: A topic proposal order based on the user's predicted gender. Based on our collected conversations and heuristic experience, we curated different topic orders for males, females, and unpredictable genders.
    \item \textbf{\texttt{used\_topics}}: A list of topics that the user has discussed or the system has proposed but was rejected by the user.
\end{itemize}
Our system would propose a topic that is in \texttt{``preferred\_topics''} if it is not in \texttt{``used\_topics''}. If all the preferred topics have been discussed, we select the highest priority topic in \texttt{``personal\_topics''} that is not in \texttt{``used\_topics''}. When all the topics have been used, we reset \texttt{``used\_topics''} and propose the first topic in \texttt{``personal\_topics''}.

\subsection{Adaptation in Topic Modules} \label{ssec:adaptation_in_topic}
Our system also adapts uniquely to each user in the topic module level.

In our Greetings Module, we first identify users by checking their user ids provided by Alexa Skill Kit. Then we use different templates to greet new users and returning users. For instance, if the user is identified as a new user, we would ask for their names; if they are a returning user, we would confirm if they are the same user by prompting for their names in case multiple users share the same device with their household. We also ask different open questions for returning users to know their recent updates (e.g., ``What have you been up to recently? '') This adaptation humanizes our system by demonstrating our ability to recognize a user that has talked to us before.

In our Fashion Module, different subtopics are proposed based on inferred user gender. For example, we will propose makeup if we predict our user is female, and propose a different subtopic such as cologne if we instead predict our user is male. In addition, we choose different words based on user-gender. For example, for the female population, we refer to fragrances as \textit{perfume}, while to the male population, we refer to fragrances as \textit{cologne}. Furthermore, we categorize similar subtopics in the same group and track how often the user agrees or disagrees within the topic. If the user repetitively dismisses our response by saying ``no'', we promptly decide to transition to a different subtopic group that the user may prefer instead of staying within the same group.
 %Yu and Kaihui
\section{Acknowledging User Utterances} \label{sec:acknowledgement}
% Sam please talk about the two methods you used for acknowledgement
Effectively signaling understanding is a key aspect of human conversation. 
According to Traum \& Hinkelman (1992) \cite{traum1992conversation}, 
human interlocutors do not assume that they have been understood until their conversational partner explicitly acknowledges understanding, through either verbal or physical actions. Since our Alexa bot is unable to nod to indicate understanding, the bot must verbally acknowledge user utterances to fulfill its obligations as a conversational agent. For example, if our bot asks a user \say{What do you like to do in your free time?} and the user responds \say{I like to dance}, it would be inappropriate for the bot to proceed with the conversation without in some way acknowledging the user's statement with a response such as \say{Ok, you like to dance.} While a simpler response such as \say{Okay!} or \say{Nice!} may be appropriate in human conversation, we feel that a more explicit response which rephrases the user input implies fuller understanding by our conversational bot. Additionally, because detecting implied sentiment in user statements can be quite challenging, our rephrase strategy allows us to avoid positive acknowledgment of a potentially negative statement. 

Similarly, we integrate a method to respond to user questions to which our bot is unable to generate a specific response. For example, if a user asks \say{What is my mother's name?}, we are able to answer \say{I don't know what your mother's name is.} Although we could simply respond \say{I don't know} to such a question, rephrasing and restating the user's query demonstrates to the user that, even if the bot is unable to answer the question, it at least understood what the user asked. According to Traum \& Hinkelman \cite{traum1992conversation}, such a response, while not answering the question directly, grounds the question and communicates that the bot has understood. Like our statement acknowledgment, this question acknowledgment method makes our bot a more engaging conversational agent.

We implement a separate system module to generate utterances to acknowledge user statements and questions. We use the method described in and code provided by Dieter et al. (2019) \cite{dieter-etal-2019-mimic} as a starting point for our acknowledgment system. Specifically, we implement a rule-based system as described in their paper, making modifications necessary to create variety and prevent unnatural repetition. We segment each of our sentences using our previously described sentence segmentation model. We then use a series of syntactic patterns and ordering rules to determine which segmented portion should be acknowledged. When necessary, we complete user utterances to resolve ellipsis using a rule-based method that considers the previous context and the syntax of the user response. For example, had the bot asked \say{What do you like to do in your free time}, and the user responded \say{Dance}, the sentence would be completed as \say{I like to dance}. Input is then parsed using the Stanford CoreNLP \cite{corenlp} constituency parser, and utterances are reworded using a set of reordering rules. When responding to a question it is unable to specifically answer, the system prepends variants of \say{I do not know}. Additionally, pronouns are replaced, and verb conjugations appropriately modified to generate a sensible and grammatical response.

In order to increase the diversity and relevance of the bot's responses, we also integrate Blender, as discussed in Section 3.5.2, to respond to certain user utterances. Blender is capable of generating meaningful responses to many user queries to which the bot does not have a specific response. For example, if the user asks \say{Have you ever heard about Roblox?}, the rule-based method described in the preceding paragraph would respond \say{I don't know if I've ever heard about Roblox.} Blender, on the other hand, will respond \say{I have heard of it, but I've never played it}, which is clearly a more engaging response. At the same time, Blender output is often generic and not topical. For example, when a user asked the system \say{Who is Maddie?}, Blender responded \say{I love movies}, conditioned on the bot's previous utterance. The rule-based system, on the other hand, would respond \say{I don't know who Maddie is.} We implement a rule-based system that determines if the Blender response contains any of the non-stop word tokens from the user input, to ensure that the Blender response is topical. If it does, we choose the Blender output over the generic rule-based acknowledgment.

In case the user intent is very common and occurs highly frequently, we use a hand-written acknowledgment template to acknowledge the user to improve user engagement instead of using the previous method to rephrase it. For example, if the user says ``I don't know'', the module generates ``That's okay''. If the user says ``That's interesting!'', it then replies, ``I'm glad you like it! ''

%Don't know if I should include this.
% Although we are currently using a rule-based acknowledgment system, our final goal is to develop a more robust method using a large pretrained neural model such as BART \cite{lewis2019bart} or DialoGPT \cite{dialogpt}. We are in the process of collecting human judgments of our system acknowledgment to fine-tune such a model for this task. % Sam

\section{Results and Analysis} \label{sec:evaluation}

\begin{figure}[htp]
\centering
\includegraphics[width=0.8\linewidth]{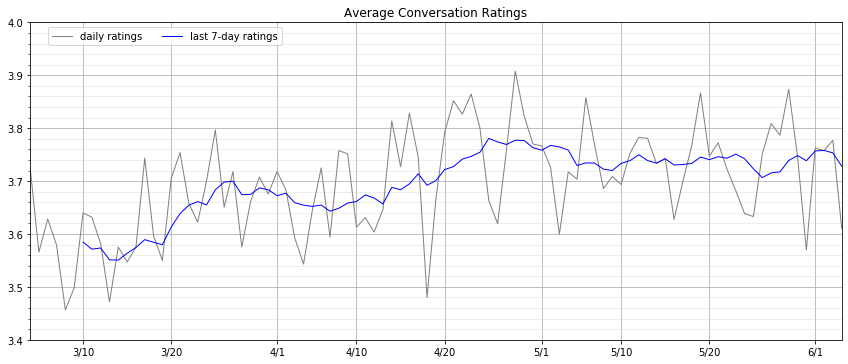}
\caption{Average user ratings (daily and last 7-day) over time}
\label{fig:rating}
\end{figure}

Figure \ref{fig:rating} shows our system's rating over time for the past three months before Final Phase ended (June \nth{4}). Our chatbot's last seven-day rating improved from $3.60$ to $3.75$ during the last two months. We reached a median conversation duration of $2:14$ (min:sec) at the end of Final Phase, and lengthened our 90-percentile conversation duration by $24\%$ (from $10:49$ to $13:31$) during the last three months. The increase in both ratings and duration reflects all the improvements we made in our system.

% We believe the main reason that led to the improvement are ...
% 
%  To further 

\subsection{Topic module performance analysis}

We use the average rating per turn to evaluate the performance of each domain-speciﬁc module, where we assume each turn contributes equally to the rating of a complete dialogue. We chose this criteria over the average rating per dialog since each module’s impact on the rating should correlate with the number of turns it hits.

\begin{figure}[hp]
\centering
\includegraphics[width=1\linewidth]{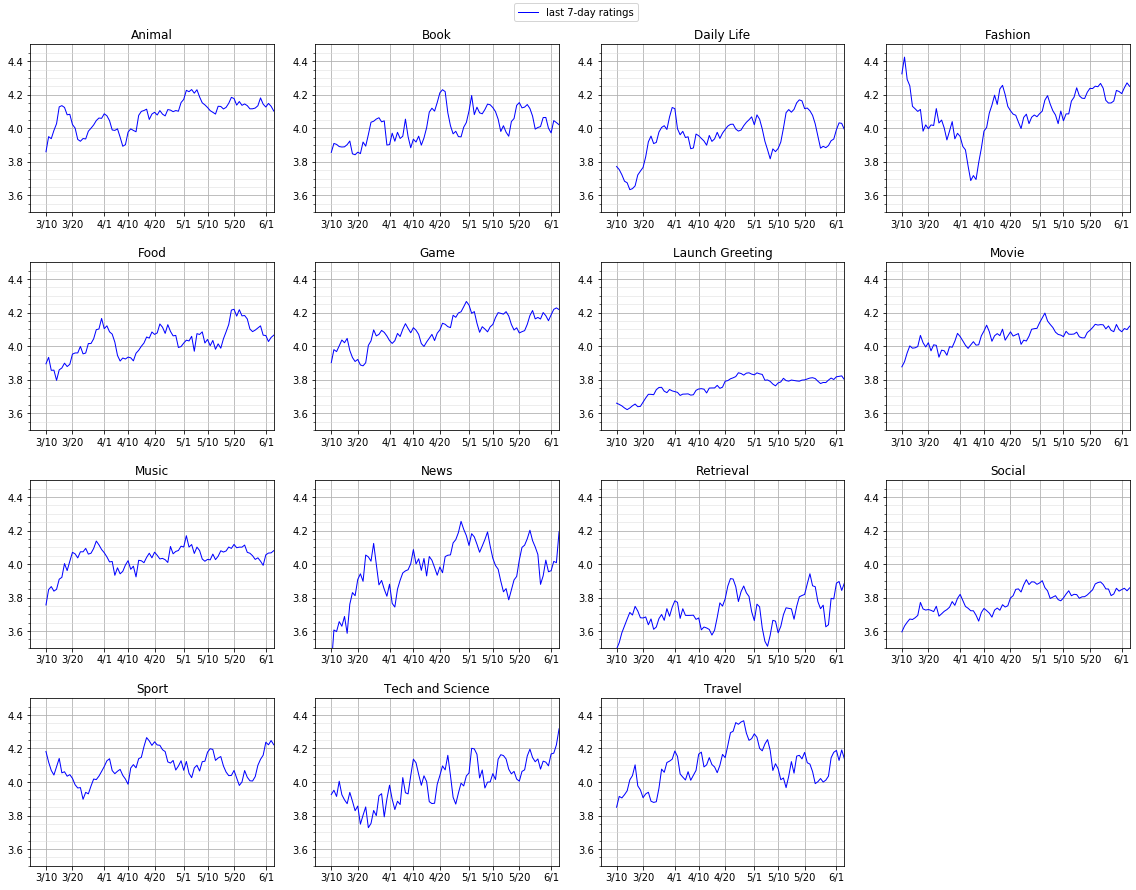}
\caption{Average user ratings per module (last 7-day) over time}
\label{fig:topic_rating_over_time}
\end{figure}

Table \ref{tab:topic_module_turn_info} demonstrates the number of turns and average ratings for each topic module. \textit{Movie} covers the most turns and has the longest average turn count per conversation, while \textit{Fashion} and \textit{Game} have the highest ratings. Figure \ref{fig:topic_rating_over_time} shows that the last 7-day rating of most modules improved over time in general. Among them, \textit{News} increased the most as we adapt the dialog flow better to corona virus-related topic.

\begin{table}[hp]
\centering
\begin{tabular}{lrrr}
\hline
\textbf{Topic Modules} &
  \multicolumn{1}{l}{\textbf{Total Turn Count}} &
  \multicolumn{1}{l}{\textbf{\begin{tabular}[c]{@{}l@{}}Average Turn Count\\ Per Conversation\end{tabular}}} &
  \multicolumn{1}{l}{Average Rating} \\ \hline
\textbf{Movie}              & 87353  & 14.51 & 4.09 \\
\textbf{Animal}             & 65672  & 12.80 & 4.13 \\
\textbf{Launch Greeting}    & 55829  & 2.92  & 3.80 \\
\textbf{Music}              & 36907  & 10.01 & 4.06 \\
\textbf{Game}               & 31456  & 8.57  & 4.16 \\
\textbf{Social}             & 28300  & 2.66  & 3.84 \\
\textbf{Sport}              & 26948  & 9.44  & 4.12 \\
\textbf{Food}               & 25117  & 10.80 & 4.09 \\
\textbf{Book}               & 18774  & 12.33 & 4.06 \\
\textbf{Tech and Science}   & 9774   & 8.08  & 4.13 \\
\textbf{News}               & 8262   & 8.12  & 4.03 \\
\textbf{Travel}             & 7877   & 8.29  & 4.09 \\
\textbf{Fashion}            & 6174   & 6.14  & 4.18 \\
\textbf{Daily Life}         & 4209   & 1.59  & 4.00 \\
\textbf{Comfort}            & 2527   & 4.84  & 3.75 \\
\textbf{Retrieval}          & 2412   & 1.86  & 3.75 \\
\textbf{Outdoor Activities} & 1044   & 2.18  & 4.02 \\
\textbf{Self-Disclosure}    & 897    & 2.49  & 2.27 \\
\textbf{Jokes}              & 560    & 3.46  & 4.19 \\ \hline
\textbf{ALL}                & 459751 & 23.98 & 4.02 \\ \hline
\end{tabular}
\caption{Overall performance of each topic module during the last month of Final ( May 4 - June 4, ordered by total turn count)}
\label{tab:topic_module_turn_info}
\end{table}

\subsection{Dialog Strategy Effectiveness} \label{sec:user_adaptation}
\subsubsection{Topic entry analysis}
\begin{figure}[hp]
\centering
\includegraphics[width=0.8\linewidth]{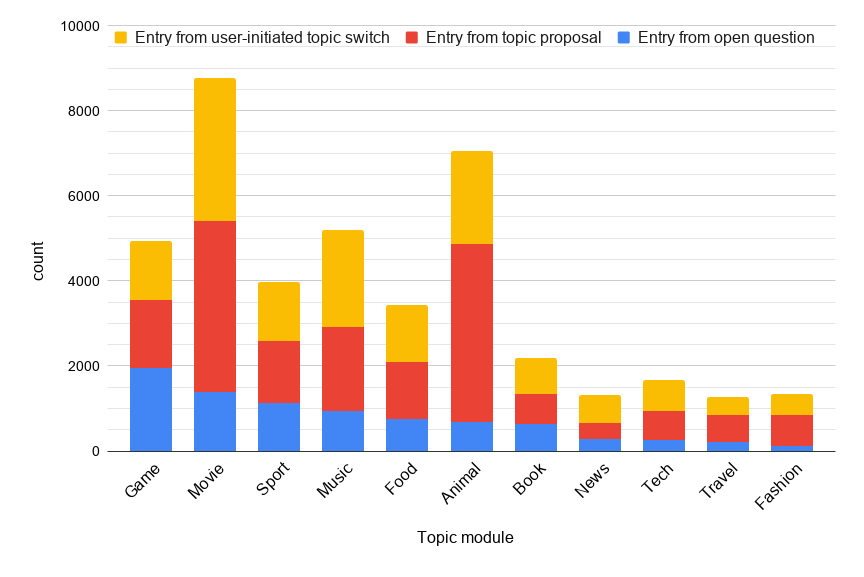}
\caption{Distribution of different entry method for each topic module during the last month of Final ( May 4 - June 4) }
\label{fig:topic_entry_distribution}
\end{figure}

% austin and yu
To understand how users enter each topic module, and evaluate if asking open questions and proposing topics are effective in helping users find topics that interest them, we calculate the distribution of the three entry ways for all modules, as shown in Figure \ref{fig:topic_entry_distribution}. Almost all modules have some entries from open questions, which means that the open questions might be effective for nailing down topics for users. It is notable that \textit{Game} and \textit{Movie (including TV)} have the highest entry count from open question, implying that users are actively interested in talking about these topics. Other than entering from open question and topic proposal, users may switch topic with a command or question (eg. ``Let's talk about dogs'' or ``Have you seen Avenger end games'') while talking about another topic. \textit{Movie}, \textit{Music} and \textit{Animal} have high number of entry from user initiated topic switch. The large number of this entry indicates that allowing users to switch to a specified topic might play an important role in maintaining user engagement.

In Table \ref{table:op_module}, we compare the performances of different entries for all topic modules. For most modules, topic proposal entries have better ratings than open question entries. It may be because user utterances are generally relatively simple in topic proposal entry (e.g., answering ``yes'' to the question ``I'm really fascinated by animals. What about you, do you like animals?'') so the following dialog flow is easier to handle and more natural. At the same time, user utterances usually are more complicated in open question entries (e.g., saying ``i like baking'' to the question ``What do you like to do for fun?''), which requires the corresponding module to detect entities and map them to the corresponding state. If the module does not have a pre-defined flow that fits that entity, the response might not meet the user's expectation and thus lead to lower ratings.

\begin{table}[htp]
\centering
\begin{tabular}{lrrr}
\hline
\textbf{Topic module} &
  \multicolumn{1}{l}{\textbf{Open question entry}} &
  \multicolumn{1}{l}{\textbf{Topic proposal entry}} &
  \multicolumn{1}{l}{\textbf{Other entry}} \\
\hline
\textbf{Movie}            & 4.05          & \textbf{4.09}          & 4.08          \\
\textbf{Book}             & 3.96          & \textbf{4.20}          & 4.03          \\
\textbf{Music}            & 4.07          & \textbf{4.10}          & 4.01          \\
\textbf{Sport}            & 4.01          & \textbf{4.20}          & 4.18          \\
\textbf{Food}             & 3.87          & \textbf{4.24}          & 4.05          \\
\textbf{Animal}           & 4.17          & 4.11                   & \textbf{4.18} \\
\textbf{Travel}           & 4.05          & \textbf{4.19}          & 3.90          \\
\textbf{Game}             & \textbf{4.22} & \textbf{4.13}          & 4.13          \\
\textbf{Fashion}          & \textbf{4.26} & 4.16                   & 4.18          \\
\textbf{Tech and Science} & 4.11          & \textbf{4.20}          & 4.16          \\
\textbf{News}             & 3.88          & \textbf{4.31}          & 3.90          \\

\hline
\end{tabular}
\caption{Rating of each topic module from different entries during the last month of Final ( May 4 - June 4)}
\label{table:op_module}
\end{table}

\subsubsection{Topic proposal analysis}
To evaluate if users are interested in the topics we proposed, we measure the user acceptance rate as displayed in Table \ref{tab:topic_acceptance_rate}. \textit{Animal} and \textit{Music} have the highest acceptance rate (0.75 and 0.72 respectively). The result indicates what topics interest the users the most, and can serve as a reference when deciding which topic modules the bot should propose first to meet most users' interests. Note that the acceptance rate does not have an evident correlation to the topic module rating. For example, \textit{Fashion} module has the lowest acceptance rate while having the highest rating as it has an engaging dialog flow. On the other hand, modules with the highest rating might not have the highest rating if the dialog does not meet the user's expectation. This reveals that dialog flow plays an important role in the bot's overall performance.

\begin{table}[]
\centering
\begin{tabular}{lrrr}
\hline
\multicolumn{1}{l}{\textbf{Topic module}} &
  \multicolumn{1}{l}{\textbf{User acceptance count}} &
  \multicolumn{1}{l}{\textbf{Topic proposal count}} &
  \multicolumn{1}{l}{\textbf{User acceptance rate}} \\ 
\hline
\textbf{Animal}           & 4192 & 5573 & 0.75 \\
\textbf{Music}            & 1976 & 2739 & 0.72 \\
\textbf{Movie}            & 4038 & 5818 & 0.69 \\
\textbf{Food}             & 1332 & 2031 & 0.66 \\
\textbf{Tech and science} & 665  & 1071 & 0.62 \\
\textbf{Travel}           & 629  & 1071 & 0.59 \\
\textbf{Book}             & 686  & 1220 & 0.56 \\
\textbf{Game}             & 1585 & 2930 & 0.54 \\
\textbf{News}             & 371  & 795  & 0.47 \\
\textbf{Sports}           & 1468 & 3220 & 0.46 \\
\textbf{Fashion}          & 727  & 1804 & 0.40 \\
\hline
\end{tabular}
\caption{Average user acceptance rate for all topic modules during the last month of Final ( May 4 - June 4)}
\label{tab:topic_acceptance_rate}
\end{table}

\subsection{Name Entity Detection}
\begin{table}[h!]
    \centering
    \begin{tabular}{l|l|l}
        \textbf{Utterance} & \textbf{Noun phrases} & \textbf{Key phrases} \\
        \hline
        what year was uhhh julius caesar murdered & uhhh julius caesar & julius caesar \\
        \hline
        can we not talk about animals left in the wild & animals, the wild, wild & animals, wild \\
        \hline
        eat at a restaurant & a restaurant & at a restaurant \\
        \hline
        tom clancy's rainbow six siege & tom clancy's, six siege & tom clancy's \\
        & & rainbow six siege \\
        \hline
        tell her wings of fire & her wings, fire & wings of fire \\
        \hline
        my dream vacation would be toronto & dream vacation & dream vacation, \\
        & & toronto \\
        \hline
        oh yeah my mom and dad have taking me & mom, dad, a bunch, & bunch, places \\
        a bunch of places & places & \\
    \hline
    \end{tabular}
    \caption{Differences between noun phrases (extracted using the constituency parser) and key phrases (output of the KeyPhrase model). Examples are cherry-picked. Key phrases are usually more precise than noun phrases.}
    \label{tab:keyphrase}
\end{table}
We evaluated the difference between the noun phrases and the KeyPhrase outputs introduced in Section~\ref{sssec:np}.
Noun phrases are extracted from the output of the constituency parser with the utterance as the input.
Key phrases are output from the KeyPhrase model.

% quantitative result - Austin

We investigated the differences between noun phrases and key phrases (Table~\ref{tab:keyphrase}).
Unlike written text, spoken utterances are noisy, so the noun phrases may include fillers.
For example, in the utterance ``what year was uhhh julius caesar murdered'', the extracted noun phrase is ``uhhh julius caesar'' instead of just ``julius caesar''.
In addition, extracting all noun phrases from a constituency parse may include overlapping phrases.
For example, both ``the wild'' and ``wild'' are extracted from the utterance ``can we not talk about animals left in the wild''.
Although including overlapping phrases may help with reasoning in topic modules, it has a negative impact on latency since there is more work that needs to be done in the entity linking step.
Additionally, key phrases capture more phrases with named entities such as ``tom clancy's rainbow six siege'', ``wings of fire'', and ``toronto''.
However, key phrases sometimes miss certain phrases such as "mom" and "dad" in the last example in Table~\ref{tab:keyphrase}.

% \begin{figure}[]
%      \centering
%      \begin{subfigure}[b]{0.4\textwidth}
%          \centering
%          \includegraphics[width=\textwidth]{Alexa Prize Technical Article LaTeX/evaluation/googlekg_ad.png}
%          \caption{Comparison between using nounphrase and keyphrase for searching google knowledge graph}
%          \label{fig:googlekg_ad}
%      \end{subfigure}
%      \begin{subfigure}[b]{0.4\textwidth}
%          \centering
%          \includegraphics[width=\textwidth]{Alexa Prize Technical Article LaTeX/evaluation/googlekg_cd.png}
%          \caption{Comparison between using keyphrase and keyphrase with similarity for searching google knowledge graph}
%          \label{fig:googlekg_cd}
%      \end{subfigure}
%     \caption{Three simple graphs}
%         \label{fig:three graphs}
% \end{figure}

% \subsubsection{Generation}
 % Austin and Kaihui
%\input{visualization}
% \import{./}{data_analytics.tex} % Sam

\section{Conclusion} \label{sec:conclusion}
We designed an open-domain social conversation system that achieved an average rating of $3.73$ out of $5$ in the last seven days of the semi-finals. We made many contributions in entity detection and linking, user-adaptive dialog management, and integrating generation and template-based methods in a social chatbot. In sum, we increased its understanding by adding more powerful neural network-based models and improved dialog response by manually creating templates with free-form generation models. These generation models handle corner cases while carefully written templates handle popular intents.  We also improved the selection of proposed topics for different users based on their profile information, such as gender and dialog act use distribution to create an adaptive and unique conversation experience for each user. %Zhou
% \import{./}{future_work} %Zhou

\subsubsection*{Acknowledgments}
We would like to acknowledge the help from Amazon in terms of financial and technical support. We would also like to thank Leann Sotelo, Michelle Cohn for helping on revising the system dialog templates; Michel Eter for providing feedback to our system.

\bibliography{main}
\bibliographystyle{acm}
\newpage

\end{document}